\documentclass[10pt,journal,compsoc]{IEEEtran}

\usepackage{times}  
\usepackage{helvet} 
\usepackage{courier}  
\usepackage[hyphens]{url}  
\usepackage{graphicx} 
\usepackage{caption} 
\usepackage{amsmath}
\usepackage{esvect}
\usepackage{xcolor}
\usepackage{soul}
\newcommand*\mean[1]{\bar{#1}}
\DeclareMathOperator*{\argmax}{arg\,max}

\ifCLASSOPTIONcompsoc
  \usepackage[nocompress]{cite}
\else
  \usepackage{cite}
\fi

\begin{document}

\title{Discovering and Interpreting Biased Concepts in Online Communities}
\author {
        Xavier Ferrer-Aran,
        Tom van Nuenen,
        Natalia Criado,
        Jose Such
\IEEEcompsocitemizethanks{\IEEEcompsocthanksitem Xavier Ferrer, Department of Informatics, King's College London. E-mail: xavier.ferrer\_aran@kcl.ac.uk}
}

\IEEEtitleabstractindextext{%
\begin{abstract}
Language carries implicit human biases, functioning both as a reflection and a perpetuation of stereotypes that people carry with them. Recently, ML-based NLP methods such as word embeddings have been shown to learn such language biases with striking accuracy. This capability of word embeddings has been successfully exploited as a tool to quantify and study human biases. However, previous studies only consider a predefined set of biased concepts to attest (e.g., whether gender is more or less associated with particular jobs), or just discover biased words without helping to understand their meaning at the conceptual level. 
As such, these approaches can be either unable to find biased concepts that have not been defined in advance, or the biases they find are difficult to interpret and study. 
This could make existing approaches unsuitable to discover and interpret biases in online communities, as such communities may carry different biases than those in mainstream culture. This paper improves upon, extends, and evaluates our previous data-driven method to automatically discover and help interpret biased concepts encoded in word embeddings. We apply this approach to study the biased concepts present in the language used in online communities and experimentally show the validity and stability of our method.
\end{abstract}

}

\maketitle
\IEEEdisplaynontitleabstractindextext

\IEEEraisesectionheading{\section{Introduction}\label{sec:introduction}}
\IEEEPARstart{L}{inguistic} biases have been the focus of human language analysis for quite some time \cite{Hamilton1986,Basow1992,Holmes2008}. Recently, it has been proven that machine learning approaches applied to corpora of human language, such as word embeddings, learn human-like semantic biases \cite{bolukbasi2016man,Caliskan2017}.\footnote{Author's copy of the manuscript published by IEEE Computer Society in journal IEEE Transactions on Knowledge \& Data Engineering \url{https://www.computer.org/csdl/journal/tk/5555/01/09667280/1zMCh7YGvfi}}
Beyond the value this has towards creating \emph{fairer} Artificial Intelligence
~\cite{barocas2017problem,zemel2013learning,criadodigital,van2020transparency,ferrer2020bias,such2017privacy}, it has also enabled compelling methods for social scientists to study human language and biases \cite{Bryson2016,Zhao2006,VanMiltenburg2016,garg2018word,kozlowski2019geometry}. That is, by looking at the biases learned by a machine learning model, it is possible to study biases in datasets in more detail. 
For instance, by training word embeddings models and analyzing them, Garg et al. \cite{garg2018word} were able to study the dynamics and the evolution of predefined biases related to gender and religion in the United States across 100 years. 

Existing approaches to study biases in word embeddings (see Section \ref{sec:related} for details), however, have one of the following limitations. That is, either: i) they are only able to attest whether arbitrarily predefined biases exist or not, so they are unable to \emph{discover} what the actual most salient biases might be in a model; or ii) they only discover biased words, which is of limited value when trying to make sense of the biases discovered and what they mean at a conceptual level. These limitations often make current approaches unsuitable to study biases in the language of online communities. 
This is because such communities often exhibit values and norms that are sometimes distinct from those held by the majority \cite{Fine1979,Marwick2017}, so anticipating potential biases in advance is difficult; and because understanding the meaning of those biases is needed to properly interpret and study them \cite{Abebe2019,tanczer2016hacktivism,creese2019stereotypes}. Studying the biases of online communities is very important to tackle the social problems they have been associated with, such as radicalization \cite{Marwick2017} and discrimination against some types of users, particularly those users with protected attributes such as gender, religion, ethnicity, social class, etc. \cite{tanczer2016hacktivism,aran2019attesting}.

In this paper, we improve upon our previous, general and data-driven approach  \cite{ferrer2020discovering} to discover  linguistic biases in word embeddings. Concepts of interest, e.g. those related to protected attributes, such as gender, religion and the like, are used to \emph{discover} the most frequent and biased words towards them (e.g. frequent words biased towards `women' rather than `men'). Differently from our previous work, in this paper: i) we improve the automatic selection of biased words, using not just the strength of biases but also their frequency; ii) we present a novel automatic categorisation of the biases, through semantic clustering and semantic analysis; and iii) we present novel methods to help interpret the resulting biased concepts, using semantic labels and rankings. All of this allows our method to help make sense of the most important biases at a conceptual level and in a more detailed manner, using the vocabulary of the online community studied.

The paper is structured as follows: Section \ref{sec:related} discusses related work and details the gap this paper addresses.
Section \ref{sec:model} introduces the notion of \emph{bias strength} in word embeddings, and the preliminaries on which our method bases. We present our approach to discover and interpreting biased concepts, together with the different bias ranking in Section \ref{sec:main}.
We apply our method to discover biased concepts in models trained using the corpora of two online communities (/r/TheRedPill, /r/Atheism) on the English-speaking discussion platform Reddit in Section \ref{sec:ev}. We provide an extensive evaluation of our method in Section \ref{sec:sb-val}, showing the stability of the biases discovered and the effect of the different parameters of our model and those used to train the embeddings model, and demonstrating the validity of our method by applying it to the general-purpose Google News pre-trained model and comparing our method with arbitrarily predefined biases in that model attested by previous work. 
Finally, we finish with some concluding remarks and pointers to the available datasets, code and demo of the tool developed based on our method in Section \ref{sec:conclusions}.

\section{Related Work} \label{sec:related}
In this section, we discuss two main streams in the most related work to this paper. One stream includes those works \emph{attesting} predefined biased concepts, and the other stream includes those works \emph{discovering} biased words in word embeddings.

Works \emph{attesting} biased concepts in word embeddings measure the association between predefined concepts usually related to known stereotypes – for instance, whether men are more often associated with a professional career while women are more often associated with family \cite{Caliskan2017,garg2018word,sutton2018biased,zhao2019gender,papakyriakopoulos2020bias,manzini2019black,kurita2019measuring, kozlowski2019geometry,brunet2018understanding}. An example of measure used to this end is the Word Embeddings Association Test (WEAT) \cite{Caliskan2017}, inspired by the Implicit Association Test (IAT) \cite{greenwald1998measuring}, widely used in psychology and the social sciences to study stereotypes \cite{kiefer2007implicit}. In particular, all the concepts involved (e.g. men/women \emph{and} career/family) are represented via sets of words, and WEAT compares distances in the word embeddings model between those sets of words using cosine similarity. 
%
While being able to attest biases between arbitrary, predefined concepts is of high value, it nonetheless requires all the involved concepts to be defined in advance, which may not be possible in online communities, as they may exhibit biases and stereotypes that are not based on those in mainstream culture \cite{Marwick2017}. In contrast, in our work, we only consider as input the \emph{attribute} concepts of interest, e.g. men/women, and our method then discovers all other concepts that are most associated with these \emph{attribute} concepts. 

Works \emph{discovering} biased words overcome some of the limitations of \emph{attesting} by enumerating all the words that are biased towards others in a word embeddings model \cite{bolukbasi2016man,zhang2018mitigating,swinger2019biases,gonen2019lipstick,bodell2019interpretable}. 
However, the resulting lists of biased words do not explain what these biases mean or to what extent they are important in the context of a community, both of which are needed to properly interpret and study the biases discovered in a community \cite{Abebe2019,tanczer2016hacktivism,creese2019stereotypes}. 
Previous work has demonstrated language innovation in the manosphere in order to understand how different groups within it view their own departure from mainstream society \cite{Farrell2020}. Socio-cultural linguistic work, which has long addressed the structure and meaning of specialised vocabularies, typically was based on smaller sets of data. The internet and social media have generated numerous communities, however, each producing vast amounts of content. Manual identification and analysis of the subcultures’ specialised vocabularies has become impractical; automatic methods are needed to help with both the identification and understanding of community language.

Our previous work in this direction is presented in \cite{ferrer2020discovering}, in which we define a system to identify a set of words biased towards certain concepts, and organize them in categories in order to compare their biases. Although the approach gives an idea of what are the most biased words towards certain concepts in a community, and to which semantic categories they belong, it has several limitations. 
First, the bias measure used does not consider the frequency when selecting the set of biased words, which means that some of the biases may be rare and unrepresentative of the community. 
Second, the threshold used to select biased words is manually determined by analyzing the distribution of bias in the community, which makes it difficult to automate. Third, the analysis of biases performed does not allow for a link between the biased concepts and the actual language used by the community. This is important as smaller communities in larger societies frequently develop specialised vocabularies that reflect unique aspects of the identity, needs and realities of group members \cite{MacQueen2001}.
A further limitation is that the work neither considered the potential impact of stability, nor did it recognize the effect of different parameters in the model.

In this paper, we combine both frequency and bias to automatically identify the most salient words in a community; that is, words that are both frequently used and strongly biased towards the two concepts of interest. Afterwards, these words are aggregated based on their semantic similarity, by clustering them using the word embeddings model.
The resulting semantically-similar clusters are analyzed in two ways. First, we categorize the major discourse field each of the clusters belongs to and provide a relative frequency of the fields. Second, clusters are ranked based on the strength of their bias, their frequency, and their sentiment to offer a more detailed view of the biases discovered and how they are expressed in the language of the community in a more or less biased, frequent, or positive/negative way. The semantic categorization and different rankings provides a context that offers both a general and more detailed view of the biases at the conceptual level over the different dimensions of bias and in an automatic manner.  

%
%
\section{Preliminaries}\label{sec:model}
The first step to be able to discover biased concepts towards the attribute concepts of interest (e.g. men/women), is to discover the specific words that are biased towards the attribute concepts. 
To do this, we need to measure the bias between words in a word embeddings model. 
Given a word embeddings model built from a text corpora, and two sets of words representing the attribute concepts (e.g. men/women) one wants to discover biases towards, a common approach in the literature\footnote{Note that there are many alternative bias definitions in the literature \cite{zhang2018mitigating, badillawefe}, such as the \emph{direct bias} measure \cite{bolukbasi2016man}, the relative norm bias metric \cite{garg2018word}, and others, all with similar results as shown by previous research~\cite{garg2018word,badillawefe}. When experimentally comparing the \emph{direct bias} metric with ours in r/TheRedPill, we obtain a Jaccard index of 0.857 (for female) and 0.864 (for male) regarding the list of 300 most-biased adjectives generated with the two bias metrics.} is to leverage the cosine similarity between embeddings to identify words close to an attribute concept (e.g. men) and far from the other attribute concept (e.g. women) as we detail below.

\textbf{Bias Strength.}
Let $T_1$ = $\{w_{0},..., w_{i}\}$ and $T_2$ = $\{w_{0}, ..., w_{j}\}$ be two sets of words that represent two different attribute concepts we want to discover biases towards --- e.g. \emph{\{she, daughter, her, ..., mother\}} and \emph{\{he, son, him, ..., father\}} describing the concepts \emph{women} and \emph{men} --- and ${c_1}$ and ${c_2}$ the centroids of $T_1$ and $T_2$ respectively, estimated by averaging the embedding vectors of all words in each attribute concept;  
we say that a word $w$ is biased towards $T_1$ with respect to $T_2$ when the cosine similarity between the embedding of ${w}$ is higher for ${c_1}$ than for ${c_2}$:

\begin{equation}
B(w, T_1, T_2) = cos({w}, {c_1} ) - cos( {w}, {c_2})
\label{eq:mb}
\end{equation}

\noindent Larger absolute values of $B$ correspond to stronger biases, positive values of bias $B$ indicate that word $w$ is biased towards $T_1$, and negative values of $B$ indicate that word $w$ is biased towards $T_2$. By applying the bias strength equation to the entire community's vocabulary, one is able to discover all the words that are biased towards attribute concepts $T_1$ with respect to $T_2$ and vice versa.

\section{Discovering and Interpreting Biased Concepts}\label{sec:main}
Beyond discovering individually biased words, as stated in the previous section, our approach aims to discover and help understand the \emph{concepts} behind the biased words, which prior work does not do. To this aim, we follow two main steps. 

In the first step, described in detail in Section \ref{sec:SSW}, we consider not only the strength of bias but also its frequency, in order to elicit the most common and frequently biased words in a community. After this, we aggregate all these resulting words into biased concepts, by clustering the most common biased words based on their embedding similarity. 

In the second step, described in detail in Section \ref{sec:genc}, we focus on facilitating the interpretation of the biased concepts. We do so in two main ways: i) by classifying every biased concept into a major discourse field, which further abstracts the meaning of the biases and facilitates a general understanding of the types of biases present in the community; and ii) by creating rankings of the biased concepts based on the strength, frequency and sentiment of the biases discovered to offer a more detailed view and nuanced analysis.

\subsection{Discovering Biased Concepts}\label{sec:SSW}

We are not only interested in the strength of the bias of the biased words, but also if biased words are commonly used, hence frequent in the vocabulary. This has two main purposes: i) to focus on the representative biases of a community \cite{Abebe2019}; and ii) to avoid instability due to low-frequency words \cite{Antoniak2018}. 

\subsubsection{Bias Salience}
Let $V$ a set for words forming the vocabulary of a word embeddings model.
We determine the salience $Sal$ of a word $w\in V$ towards an attribute set $T_1$ w.r.t an attribute set $T_2$, by combining the normalized bias from Equation \ref{eq:mb} with the normalized frequency rank\footnote{The reason to use the frequency rank instead of raw frequencies is that, since word frequencies are known to follow the Zipf's law \cite{newman2005power}, the difference in frequency is large between the most frequent words, which could then skew saliency towards frequency rather than having a balance between strength and frequency of the biases. Therefore, we use the word frequency rank instead of raw frequency to adequately smooth its importance in the equation. Note that this also ensures that when we select the most salient words later, we discard those that may have very similar frequencies but different ranks at the tail of the frequency distribution.} of the word $w$ in the corpus---denoted by $R(w) $ and assigning the most frequent word in $V$ rank 1 and the least frequent word rank $|V|$.
Words with higher values of salience $Sal \in [0,1]$ will be both frequent and biased towards the attribute set $T_1$ with respect to the attribute set $T_2$. Therefore, salience is computed as follows:

\begin{equation}
Sal(w, T_1, T_2) = \left(1- \frac{R(w)-1}{ |V|-1 }\right) \cdot 
  \left(\frac{B(w, T_1, T_2)}{\displaystyle \max_{w_j\in V} B(w_j, T_1, T_2) }\right)
\label{eq:salience}
\end{equation}\\

Even when knowing the strength and frequency of biased words, considering each of them as a separate unit is not enough to analyse and understand biases at a more conceptual level. There is a need to semantically combine related terms under broader rubrics in order to facilitate the comprehension of the biases. We start by selecting the most salient words, which are then aggregated through k-means clustering on the word embeddings. 

\subsubsection{Most Salient Words}
We order all $w \in V$ by salience, using Equation \ref{eq:salience}, towards an attribute set $T_1$ with respect to an attribute set $T_2$ (e.g. the set of words representing women with respect to the set for words representing men), and select the most salient words to focus on the most prominent biases. 

In order to do this, we consider the distribution of salience values for all the words in the vocabulary $V$ (denoted by  $X$) and we consider $n$ standard deviations plus the mean of the salience distribution ($n*\sigma(X) + \mean{X}$) as the threshold to select the most salient words --- we show later, in the extensive experiments conducted and described in Section~\ref{sec:parameters}, a characterization of how our method behaves with different $n$. That is, we select words biased towards each attribute set with salience greater or equal than $\delta =n*\sigma(X) + \mean{X}$. All the words in the vocabulary $V$ with salience towards $T_1$ with respect to $T_2$ higher than a threshold $\delta$ form the set $S_1$. In a similar manner, we create the set of salient words $S_2$ by considering salience towards $T_2$ with respect to $T_1$. These words are then used for the semantic clustering, as detailed below. 

\subsubsection{Semantic Word Clustering} 
For each set of the most salient words in $S_1$ and $S_2$ and their embeddings, we use k-means to aggregate the most semantically similar words into clusters, based on the distance between word vectors.
For each set of word vectors associated with the most salient words (denoted by $S_1$ and $S_2$), we apply k-means setting the number of clusters to all values within the $ [2,\ldots,|V|]$ interval, and we select the partition that maximizes the silhouette score~\cite{rousseeuw1987silhouettes}.\footnote{We use silhouette for its simplicity and visual clues, but our method is agnostic to the clustering algorithm and other alternatives, such as \cite{hamerly2004learning,garg2018supervising}, could also be used in conjunction with our method.} We repeat the clustering $\tau$ times in order to obtain the best partition by considering different k-means random initialisation variables ---we show later experimentally how our method behaves with different $\tau$. Notice that salient words $S_1$ and $S_2$ are clustered per separate hence resulting in two different partitions, $K_1$ and $K_2$ respectively.

After clustering, as we then have all the sets of words needed, we apply WEAT \cite{Caliskan2017}
between all clusters from both partitions $K_1$ and $K_2$ and attribute concepts $ T_1$ and $T_2$, and only keep these clusters of each partition that return significant p-values\footnote{We correct for multiple tests to reduce Type I errors using Benjamini-Hochberg (BH) with 10\% false discovery rate~\cite{benjamini1995controlling}. BH is considered more powerful than other too conservative methods like Bonferroni, which consider all p-values equally, leading to Type II errors. BH corrects p-values according to their ranking, with a better trade-off between Type I and II errors~\cite{diz2011multiple}.} towards each respective attribute concept when compared with all clusters from the other partition. In this way, we make sure that the biased concepts represented in every cluster are relevant and strongly associated to each attribute concept. 

The resulting sets of clusters for partitions $K_1$ and $K_2$ represent the most salient and representative biased concepts towards the attribute concepts $T_1$ and $T_2$ in the community.

\subsection{Interpreting Biased Concepts}\label{sec:genc}
In this section, we focus on obtaining a general understanding of the clusters found in a given partition $K$ to facilitate the interpretation of the discovered biased concepts. Since the most salient words are grouped into semantically meaningful clusters, here we aim to organize and understand the meaning of the different clusters  and facilitate the comparison between the biases towards each attribute concept. 
We do so in two main ways: by first categorizing the clusters via semantic tagging and analyzing the frequencies of the semantic tags, and second, by ranking the clusters of a partition based on different metrics. By combining these two ways, our method is able to offer both a \emph{general} and a \emph{detailed} view of the biased concepts of a community.

\subsubsection{Categorising Biased Concepts}
To give an overview of the biased concepts in a community, we conduct a semantic analysis. In particular, we tag every cluster in a partition with the most frequent semantic fields (domains) among its words. That is, for each cluster $C=\{w_0,\ldots,w_n\}$ in a partition $K$, we first obtain the semantic domain associated with each word $w\in C$ (denoted by $s(w)\subseteq D$, where $D$ is the set of all semantic domains). We also obtain the multiset of semantic domains associated with the cluster as the sum of the semantic domains associated with its words, i.e. $SC=[s(w_0)] +\ldots+[s(w_n)]$, and then we select the semantic domain that is most frequent among all the words in the cluster as the cluster tag, i.e. $TC=\argmax_{d\in D}m_{SC}(d)$, where $m_{SC}(d)$ denotes the multiplicity of semantic domain $d$ in the multiset of semantic domains associated with cluster $C$. Regarding the specific semantic domains considered, we base on the UCREL Semantic Analysis System (USAS) \cite{rayson2004ucrel}, which has a multi-tier structure with 21 major discourse fields subdivided in more fine-grained semantic domains\footnote{\url{http://ucrel.lancs.ac.uk/usas/USASSemanticTagset.pdf}} such as \emph{People}, \emph{Relationships}, \emph{Power}, \emph{Ethics} and it has been extensively and successfully used for many tasks, such as the automatic content analysis of discourses \cite{wilson1993automatic} and as a translator assistant \cite{sharoff2006assist}.

After the semantic tagging, we then analyze the frequency of the different semantic tags in a given partition $K=\{C_1,\ldots,C_n\}$. In particular, we compute the multiset formed by the tag of each cluster in the partition, denoted by $TK=[TC_1]+\ldots+[TC_n]$,  and calculate the relative frequency of each semantic domain $d\in D$ as $r_d=\frac{m_{TK}(d)}{|TK|}$, where $|TK|$ is the total number of semantic domain tags in a partition $K$, and $m_{TK}(d)$ is the multiplicity of the semantic domain $d$ in the multiset $TK$. This allows us to get a general idea of the biased concepts in a partition. For instance, this could show that for a given partition $K$ (e.g. the clusters of the most salient biases towards men), the relative frequency of \emph{Power} ($r_{Power}$) is higher than the relative frequency of \emph{Relationships} ($r_{Relationships}$) --- as we, in fact, see in one of the datasets (/r/TheRedPill) explored in the application of our method to Reddit communities later on, as described in Section \ref{sec:ev}.

\subsubsection{Ranking Biased Concepts}\label{sec:rc}
While showing the relative frequencies of the categories of the biased concepts found is informative as an abstract, a birds-eye view of the biases in a community, the distinctions marked may be coarse-grained \cite{rayson2008key}. 
Following the example above, even though \emph{Power} may be a frequent category, we do not actually know how this is expressed in the language of the community, nor how strongly or frequently biased these clusters are towards men. It also does not indicate whether the biases are typically expressed in a positive or negative way. 

In this section, we present a more detailed analysis of the biased concepts found. Instead of looking at an aggregate view of categories of clusters as in the previous section, we focus on establishing different methods to rank each of the clusters discovered. In particular, we define three different metrics to prioritise the clusters of a partition $K$ considering: i) the frequency of the bias; ii) the strength of the bias; iii) and the sentiment polarity of the bias. The rankings offer three different but complementary views to help understand the biases found in the community. 

We define the rankings based on the clusters of each partition. In particular, given a cluster $C$ in a partition $K$, we define the following metrics:

\begin{enumerate}
    \item $R_{f}$, which measures the aggregated frequency in the text corpus of the words within cluster $C$,  therefore establishing a ranking of the most common biases:
    
    \begin{equation}
        R_{f}(C) = \sum_{w\in C} F(w)
    \label{eq:rfreq}
    \end{equation}
    
    \noindent where $F(w)$ is the frequency of word $w$ in the text corpus.\\
    
    \item $R_{e}$, which orders the clusters based on the average bias strength of the words in the cluster with respect to the attribute concepts $T_1$ and $T_2$ (see Equation \ref{eq:mb}). This method assigns higher scores to the clusters that are more biased, and it is useful to identify relevant biases towards an attribute concept when compared to another:
    \begin{equation}
        R_{e}(C, T_1, T_2) = \sum_{w\in C} B(w, T_1, T_2) \cdot \frac{1}{ |C| }    
    \label{eq:rbias}
    \end{equation}
    
    \noindent where $|C|$ is the size of the cluster.\\
    
    \item $R_{s}$, which orders the clusters based on the average sentiment of the words in the cluster. We particularly consider rankings of both the most positive and most negative biases. Although strong negative polarities might be indicative of perilous biases towards a specific population, the fact that a cluster/word is not tagged with a negative sentiment does not exclude it from being discriminatory in certain contexts. We particularly define $R_{s}$ as follows:
    
    \begin{equation}
        R_{s}(C) = \sum_{w\in C} SA(w) \cdot \frac{1}{ |C| }
    \label{eq:rsent}
    \end{equation}
    \noindent where $SA$ returns a value $\in [-1, 1]$ corresponding to the sentiment polarity of a word $w$, with -1 being strongly negative and +1 strongly positive. Note that our model is agnostic to the sentiment analysis model used and different tools may be used \cite{feldman2013techniques}. 
    
\end{enumerate}

%
%

\section{Implementation on Reddit}\label{sec:ev}
In this section, we use our method to discover biases in textual corpora collected from two Reddit communities, /r/TheRedPill and /r/Atheism.

Although both communities are suspected to incorporate significant gender and religion biases respectively~\cite{Watson2016}, less is known about the form these biases take -- e.g., what the concepts are that are more biased towards women than towards men. Table \ref{tb:datasets} summarizes the datasets, including the protected attribute (\emph{P.Attr}) we discover biases towards, 
the quantity of unique \emph{Authors}, \emph{Comments} and \emph{Words}, words per comment (\emph{Wpc}), and vocabulary \emph{Density} (ratio of unique words to total number of words). The sets of words representing the attribute concepts gender and religion are taken from previous work \cite{garg2018word,nosek2002harvesting} and listed in Appendix~\ref{ap:targetsets}.  

\setlength{\tabcolsep}{2.2pt}
\begin{table}[ht]
\centering
\caption{Datasets of Reddit communities used in this paper.}
\scriptsize
\begin{tabular}{|l|lllllll|}
\hline
Dataset & P.Attr & Years           & Authors    & Comments & Words & Wpc & Density\\ \hline
/r/TheRedPill    &  gender & 2012-18      & 106,161    &  2,844,130    & 59,712  & 52.58   & $3.99\cdot10^{-4}$\\        
/r/Atheism       &  religion & 2008-09    & 699,994    & 8,668,991     & 81,114 & 38.27    & $2.44\cdot10^{-4} $\\  \hline     
\end{tabular}
\label{tb:datasets}
\end{table}

The datasets were collected using the Pushshift data platform \cite{Baumgartner2020}. The two Reddit models were trained using an Intel i5-9600K @3.70GHz with 32GB RAM, and an NVIDIA TitanXP GPU. 
To create an embedding model for each corpus, we first preprocess each comment by removing special characters, splitting text into sentences, and transforming all words to lowercase. 
Then, for each subreddit, we train a skip-gram word2vec\footnote{Using word2vec allows us to validate our method against the widely-used word2vec Google News pre-trained model (Section~\ref{sec:sb-val}). However, our method can easily be extended to other models like ELMo~\cite{peters2018deep} or BERT~\cite{devlin2018bert}.} word embeddings model, using embeddings of 200 dimensions, words with at least 10 occurrences, a 4-word window and 100 epochs as training parameters, and $n=4$ (four standard deviations) and $\tau=200$ repetitions as parameters for our method. In Section \ref{sec:parameters}, we offer an extensive analysis varying all these parameters, which shows similar results at the conceptual level regardless of the parameter choice.
For the sentiment polarity, we used the \texttt{nltk} sentiment analysis system~\cite{hutto2014vader}.
All our code is available publicly (see Section \ref{sec:conclusions}).

\subsection{/r/TheRedPill}\label{sec:trp}

\begin{figure}[h]
\centering
  \vspace{-5pt}
  \includegraphics[width=1\linewidth]{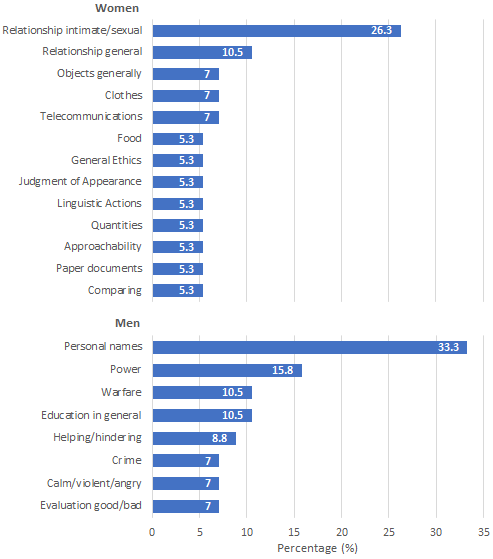}
  \caption{Relative frequency of semantic categories for biased concepts towards \emph{women} (top) and \emph{men} (bottom) in /r/TheRedPill.}
  \label{fig:trp_usastrp}
\end{figure}

\begin{figure*}[h]
    \centering
  \includegraphics[width=.8\linewidth]{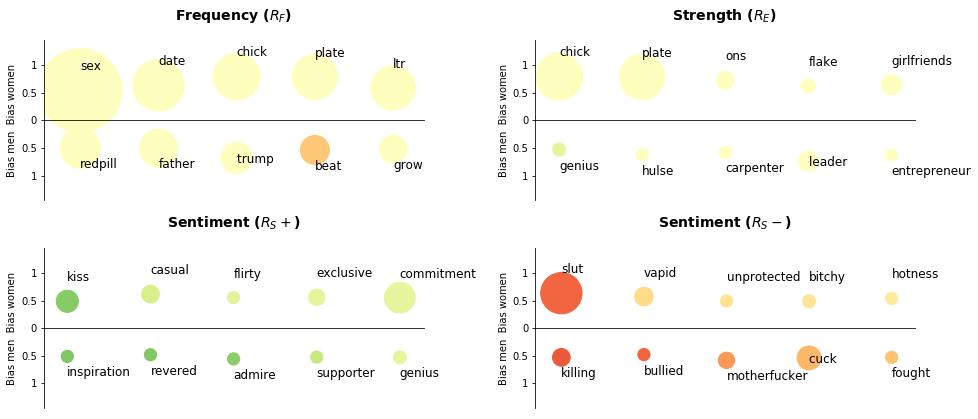}
  \vspace{-8pt}

  \caption{Top-5 clusters, labelled with their most frequent word, biased towards \emph{women} and \emph{men} in /r/TheRedPill, ranked by most frequent ($R_F$), strong ($R_E$), sentimentally positive ($R_{S}+$), and sentimentally negative ($R_S$-) bias.  Higher ranking clusters are shown ranked from left to right (x-axis), the area of each cluster corresponds to its frequency (quantity of times the words of the clusters were found in the dataset), the color with its average sentiment, and the y-axis shows the average salience of the words in the cluster. Finally, each of the clusters is tagged with its most frequent word.
  }
  \label{fig:trp}

\end{figure*}

The /r/TheRedPill community defines itself as a forum for the `discussion of sexual strategy in a culture increasingly lacking a positive identity for men’ \cite{Watson2016}. The name of r/TheRedPill is a reference to the 1999 film The Matrix: `swallowing the red pill,’ in the community’s parlance, signals the acceptance of an alternative social framework in which men, not women, have been structurally disenfranchised in the west.
In response, men must protect themselves against a ‘misandrist’ culture and the feminising of society \cite{Marwick2017,LaViolette2019}. More generally, the `red pill' construct, found across many communities in the Manosphere, is an umbrella term for a set of beliefs that men and women are categorically different on the basis of a perception of clearly defined sex \cite{Schmitz2016}.
Communities such as The Red Pill are related to the online Manosphere \cite{Ging2017,Mountford2018}, a term used to describe a collection of predominantly web-based misogynist ideologies, which include deterministic views of masculinity and femininity \cite{Schmitz2016}.
Previous research shows that the Manosphere is characterized by a volume of hateful speech that is significantly higher than other Web communities \cite{Ribeiro2020b}, and we expect these biases to be present.
We applied our method to /r/TheRedPill in order to be able to discover the exact biased concepts related to gender present in the community. More interestingly, our method enables the study of the shape these biases take, i.e., how these biases are actually expressed in the community's language. This is relevant as fringe communities such as these tend to produce concepts and expressions that are not in common use and represent the new realities, norms and values of subcommunities \cite{Farrell2020}.

After applying our method to /r/TheRedPill to discover biases towards women and men considering nouns, adjectives and verbs, we obtain the most salient words towards both attribute concepts, with sizes 216 and 194 respectively. These were then clustered into 93 and 102 clusters, with a maximum of 10 and 8 words, a minimum of 1, and a mean of 2.32 and 1.90 words with a standard deviation of 1.78 and 1.33 for women and men, respectively. The WEATs performed were all with p-values between $1.2 \cdot10^{-3}$ and $1.5 \cdot10^{-2}$, passing Benjamini-Hochberg correction for multiple tests as introduced before.

Figure \ref{fig:trp_usastrp} shows the distribution\footnote{Note that for the sake of clarity and readability,
only categories tagging at least 2\% of the clusters are shown in the distribution.} of the categories among clusters for \emph{women} and \emph{men}. 
The most frequent biases towards \emph{women} refer to relationships (and particularly \emph{sexual relationships}, used to tag the 26.3\% of the total number of biased concepts discovered), appearance (Clothes and Judgement of Appearance, adding up to 12.3\%) and objects (Objects generally, 7\%). Following, the most frequent biases towards \emph{men} refer to Personal Names (adding up to 33\% of the conceptual clusters biased towards men, discussed below in the detailed analysis), Power (15.8\%), Warfare (10.5\%) and Violence/Crime (adding up to 14\%, including the categories Calm/violent/angry and Crime). 
The semantic categorization of discovered biases shows a very different picture of the biases found towards the two genders: the most frequent semantic labels used to tag biased concepts towards women are predominantly related to objectification, i.e., appearance and sex, while men are predominantly described in relation to positions of power and strength, i.e., becoming successful agential subjects in the realm of dating and sex.

A more detailed view is shown in 
Figure \ref{fig:trp}, which compares the top-5 clusters for women and men in /r/TheRedPill, labeled using the most frequent word in the cluster, and ranked by the frequency, strength, and sentiment of the bias. In each figure, clusters are shown ranked from left to right starting with the highest ranked cluster and ignoring clusters with the same stem to show different biased concepts. The top part of the figure presents the biases towards women, while the biased concept towards men are shown on the bottom.
The area of each cluster corresponds to  the aggregated frequency of its words, that is, the quantity of times the words of the clusters were found in the subreddit. For instance, in r/TheRedPill, cluster `sex' the most frequent biased concept for women while `redpill' is the most frequent biased concept for men. Cluster colors correspond with its average sentiment, ranging from negative (red), to neutral (yellow), to positive (green), and the y-axis shows the average salience of the words in the cluster.

The most-frequent and strongest biases towards \emph{women} appear in clusters labeled `sex' or `slut', which signal obvious biased language towards women through objectification. We can also see clusters with particular jargon such as `ons' (one night stand), `chick', `plate' (a man or woman who is used for the purpose of sex),  or `flirty'. In the positively charged clusters, the appearance of the terms `commitment' and `exclusive' refers to issues of promiscuity and the ``evolutionary'' explanations for women’s loyalty that previous studies have connected to the Red Pill community \cite{Mountford2018}.

For \emph{men}, the picture is quite different. The top clusters biased towards men stand out due to their focus on strength, power and violence, concurring with the general categories identified previously. Many of the men-biased top clusters, such as `inspiration', `genius', `leader', and even `trump' (found in the same cluster with other  personal names associated with leadership, such as `obama'), represent role models related with strength and power, while others clearly allude to violence like `killing', `beat', etc. 
Finally, the most frequent negative cluster for men contains the term `cuck', signalling a 
conflict of masculinity identified in previous research, in which men struggle against one another for access to power within the dominant group \cite{Ging2017,Farrell2020}.

Our findings – i.e., that women are objectified whilst men engage in articulations of aggrieved manhood – are in line with and confirm previous social science studies on /r/TheRedPill regarding the hostile, misogynistic language in this community \cite{Ging2017,Mountford2018,Farrell2020}. The fact that our models have isolated these issues indicates their reliability in replicating existing theory. It suggests that our method finds relevant biased concepts, and that it enables an exploration of the relations between words and concepts, as constructed by the community itself. We further evaluate the quantitative validity of our method in Section \ref{sec:val}.

\subsection{/r/Atheism}\label{sec:ath}
The /r/Atheism subreddit is a large community that calls itself `the web's largest atheist forum', on which `[a]ll topics related to atheism, agnosticism and secular living are welcome'. Despite these goals, reading many of the comments on the subreddit leads to suspicions about biases towards different religions – particularly Islam vs Christianity – although these biases seem less evident than in the r/TheRedPill example.
After applying our method considering nouns, adjectives and verbs, to /r/Atheism to discover biases towards \emph{Islam} and \emph{Christianity} (the two largest religions), we obtained the most salient words towards both attribute concepts, with sizes 516 and 381 respectively. These were then clustered into 188 and 178 clusters, with a maximum of 19 and 17 words, a minimum of 1 word, and a mean of 2.74 and 2.14 words per cluster with a standard deviation of 3.03 and 2.10 for \emph{Islam} and \emph{Christianity}, respectively. The WEATs performed were all with p-values ranging from $2.5\cdot10^{-7}$ to  $3\cdot10^{-4}$, and passed  Benjamini-Hochberg correction test as introduced before.

\begin{figure}[h]
  \centering
   \vspace{-7pt}
  \includegraphics[width=.9\linewidth]{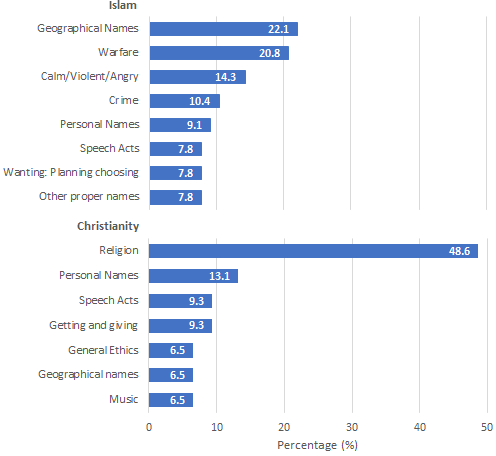}
  \caption{Relative frequency of semantic categories for biased concepts towards \emph{Islam} (top) and \emph{Christianity} (bottom) in /r/Atheism.}
  \label{fig:trp_usasislam}
\end{figure}

Figure \ref{fig:trp_usasislam} shows the distribution 
of the semantic categories among biased concepts for \emph{Islam} and \emph{Christianity}.
The figure shows a clear difference between the two attribute concepts \emph{Islam} and \emph{Christianity}. The biased concepts for Islam are categorised in normative terms such as Warfare (20.8\%), Calm/Violent/Angry (14.3\%), and Crime (10.4\%), together with names (aggregating Geographical names, Personal names and Other proper names, and adding up to 39\% of all biased concepts for \emph{Islam}; more on this later in the detailed analysis). 
Clusters about Christianity, on the other hand, fall mainly under the more descriptive and generic category of Religion (with almost half of the biased concepts for \emph{Christianity} being tagged with this semantic label), followed by Personal names (13.1\%), Speech acts (9.3\%) and General Ethics (6.5\%). These broad groupings suggest a difference in evaluative orientation towards the two religions.

\begin{figure*}[h]
\centering
  \includegraphics[width=.8\linewidth]{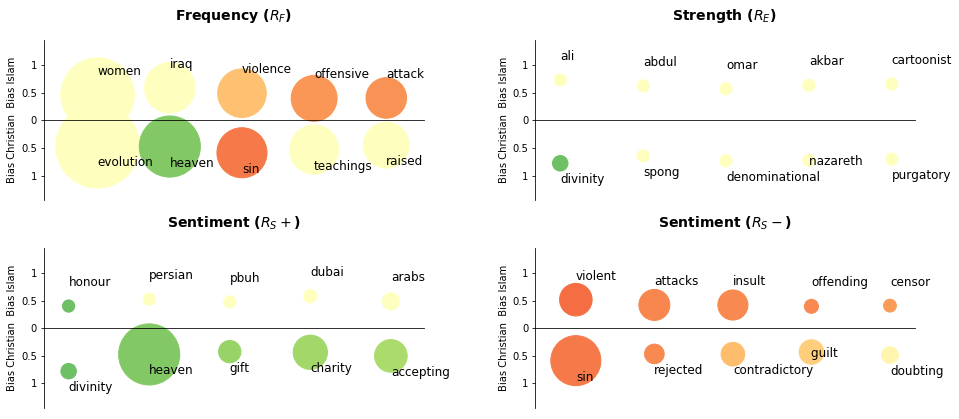}
  \vspace{-8pt}
  \caption{Top-5 clusters, labelled with their most frequent word, biased towards \emph{Islam} and \emph{Christianity} in /r/Atheism, ranked by most frequent ($R_F$), strong ($R_E$), sentimentally positive ($R_{S}+$), and sentimentally negative ($R_S$-) bias.}
  \label{fig:ath}
\end{figure*}

A more detailed view is shown in Figure \ref{fig:ath}, which compares the top-5 clusters for \emph{Islam} and \emph{Christianity}, ranked with the different rankings defined in Section \ref{sec:rc}. In each figure, Islam-biased clusters are shown on top and Christian-biased clusters on the bottom, the y-axis corresponds with average salience of the cluster, size corresponds with frequency (cluster `evolution' is the most frequent in /r/Atheism, with more than 259K hits), and color with average sentiment.
Observing the top most frequent biased clusters for both religions in the frequency ranking ($R_F$), the most frequent clusters biased towards \emph{Christianity} contain general doctrinal concepts (e.g. `heaven', `sin', `teachings'). 
Half of the top 5 most frequent clusters biased towards \emph{Islam}, however, are related to violence (`violence', `offensive', `attack').
Moreover, the `violence' cluster contains other strongly stereotyped terms such as `terrorism', `jihad', and `extremism', suggesting anti-Muslim sentiments and Islamophobia. Further proof of this would need to be further substantiated through closer inspection of the context of these terms, but our approach does produce striking differences. 

When we look at the strongest-biased clusters, we see that personal nouns describe the majority of the clusters biased towards \emph{Islam}, such as `ali', `abdul', and `omar'. These refer to public figures associated with socio-political issues of \emph{Islam}. The cluster `ali' also includes the terms `hirsi' and `ayaan', thus referring to activist Ayaan Hirsi Ali, who is known for her critical stance on \emph{Islam} and practices such as forced marriage and honor violence. The cluster `omar' also includes the terms `sheikh' and `ahmed', which likely refer to Ahmed Omar Saeed Sheikh, a British militant who was found guilty for the 1994 kidnappings of Western tourists in India. These names, then, signal a similar concern for issues of hostility and violence we saw in the previous ranking. For \emph{Christianity}, most of the clusters again refer to more generic concepts about religion such as `divinity', or `nazareth'. This could indicate that the discourse on /r/Atheism, when dealing with \emph{Islam}, revolves more around particular people and the news events they appear in, whereas \emph{Christian} discourse is less topical and political, and revolves around theological concepts and concerns.

Finally, the distribution of sentiment across all clusters biased towards \emph{Islam} is clearly more negative than \emph{Christian}-biased clusters, and gives an idea on some of the most common biased concepts associated with both religions in /r/Atheism. The most negative clusters biased towards \emph{Islam} are relatively frequent, biased and closely related (again) with violence, such as `violent', `attacks' or `insult'. On top of that, we only find one positive cluster (`honour') for \emph{Islam} (shown in $R_{s}+$ ranking), in contrast to the various positive clusters biased towards \emph{Christianity}.
Clustered words biased towards \emph{Christianity} are slightly more positive, and refer to religious terms of reward or punishment,  such as `divinity', `heaven', `gift' and `sin'. 
In general, the results suggest broad socio-cultural perceptions and stereotypes that characterize the discourse in /r/atheism community and that frequently associate Islam-biased clusters to negative connotations in contrast to Christian-biased concepts.

\section{Evaluation}\label{sec:sb-val}

In this section, we provide an extensive evaluation of our method considering different dimensions: the stability of the biases discovered, the influence the model and learning parameters may have on them, and the validity of the biases discovered. 
We particularly analyze the stability of our models in Section~\ref{sec:st}, and the importance of the different parameters used for our method and when training the embedding models in Section~\ref{sec:parameters}. We finally evaluate our approach on discovering biased concepts biases in online communities in Section \ref{sec:val}, by comparing the biased concepts discovered by our method with the arbitrary, predefined biases attested by previous work in the Google News pre-trained model.

\subsection{Stability}\label{sec:st}
Word embeddings may be unstable, particularly when trained with smaller corpora, so that small changes during training (e.g. new data) may result in different vector descriptions for the same word \cite{Antoniak2018,shiebler2018fighting}. This could be a problem, 
since different vector descriptions could result in different sets of biased concepts discovered and influence the analysis of biases of the online community. 
Although our method provides some stability mechanisms, such as the frequency of words and semantic aggregation into clusters, it is based on the same vector descriptions. 

We tested the stability of the biases found in Sections \ref{sec:trp} \& \ref{sec:ath}, following \cite{Antoniak2018}, by training four new embedding models for each of the two datasets explored,  randomly selecting 50\% of the comments of the original models to reduce the training corpora (increasing the chance for instability), 
and using the same preprocessing and parameters as before. 
This resulted in eight new models trained with a random half of the comments: $TRP_1$ -- $TRP_4$ for /r/TheRedPill, and $ATH_1$ -- $ATH_4$ for /r/Atheism. For each of the new models
we applied our method as before. 

\begin{figure}
\centering
   \includegraphics[width=0.6\linewidth]{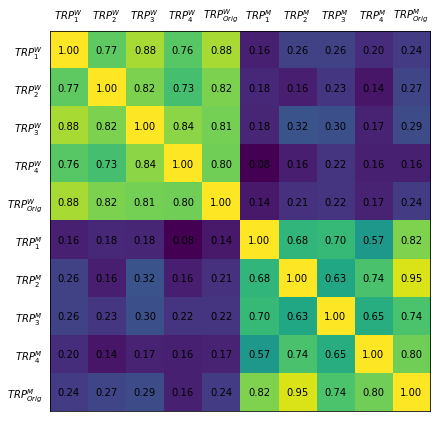}
 \vspace{-6pt}
  \caption{Overlap coefficient between bootstrapped $TRP_{1-4}$ and original $TRP_{Orig}$ models for  women ($W$) and men ($M$).}
  \label{fig:trpsimusas}
\end{figure}

Figure \ref{fig:trpsimusas} shows the overlap coefficient~\cite{vijaymeena2016survey}, also known as Szymkiewicz–Simpson, of the sets of semantic categories for the 4 new models for /r/TheRedPill ($TRP_1$ to $TRP_4$) and its original model ($TRP_{Orig}$), per attribute concepts women $W$ and men $M$ (shown as superscripts), used by more than 1\% of the clusters to improve readability. 
The higher the overlap, the closer to one, the lower the overlap, the closer to zero. 
The figure shows that the overlap between \emph{women-} and \emph{men}-biased semantic tags is small, meaning that there is a clear difference between the semantic categories used to label the concepts biased towards both attribute concepts in all models.
In addition, and when considering women and men attribute concepts per separate, the set of most frequent tags have an average overlap of 0.83 for both \emph{women} and \emph{men} with the original model $TRP_{Orig}$. That indicates that a very similar set of semantic tags was used to label women-biased concepts (and men-biased concepts, separately) across all models.

Finally, 
the set of biases towards the same attribute concept are very similar in all new models ($TRP_{1...4}$).

\begin{figure}[ht!]
\centering
  \includegraphics[width=0.6\linewidth]{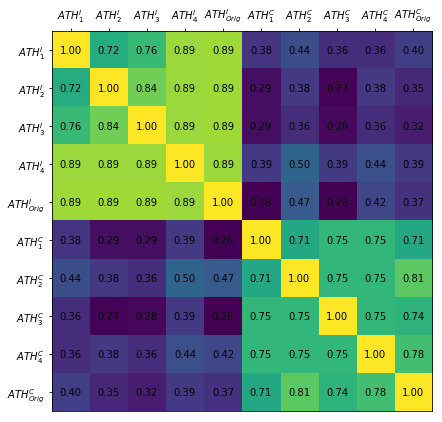}
 \vspace{-6pt}
  \caption{Overlap coefficient between bootstrapped $ATH_{1-4}$ and original $ATH_{Orig}$ models for \emph{Islam} ($I$) and \emph{Christianity} ($C$).}
  \label{fig:ap_ath}
\end{figure}

Figure \ref{fig:ap_ath} shows the overlap coefficient with the same colour code as in the previous figure for the 4 new models for /r/atheism ($ATH_{1}$ to $ATH_{4}$) and its original model($ATH_{Orig}$), per attribute concept  \emph{Islam} ($I$) and \emph{Christian} ($C$) (shown as superscripts), used by more than 1\% of the clusters to improve readability. 
The figure shows that the overlap between Islam and Christian biased labels is small, keeping the differences between them across models. Also, the overlap coefficient between  biases towards the same attribute concept is large, indicating that our approach is able to consistently identify similar biased concepts towards Islam and Christian across models in /r/Atheism. Therefore, given the results obtained, we can conclude that our method is able to pick up consistent biases, mitigating many stability issues associated with word embeddings.

\subsection{Parameter Influence}\label{sec:parameters}
\begin{figure}[ht]
\centering
  \includegraphics[width=0.80\linewidth]{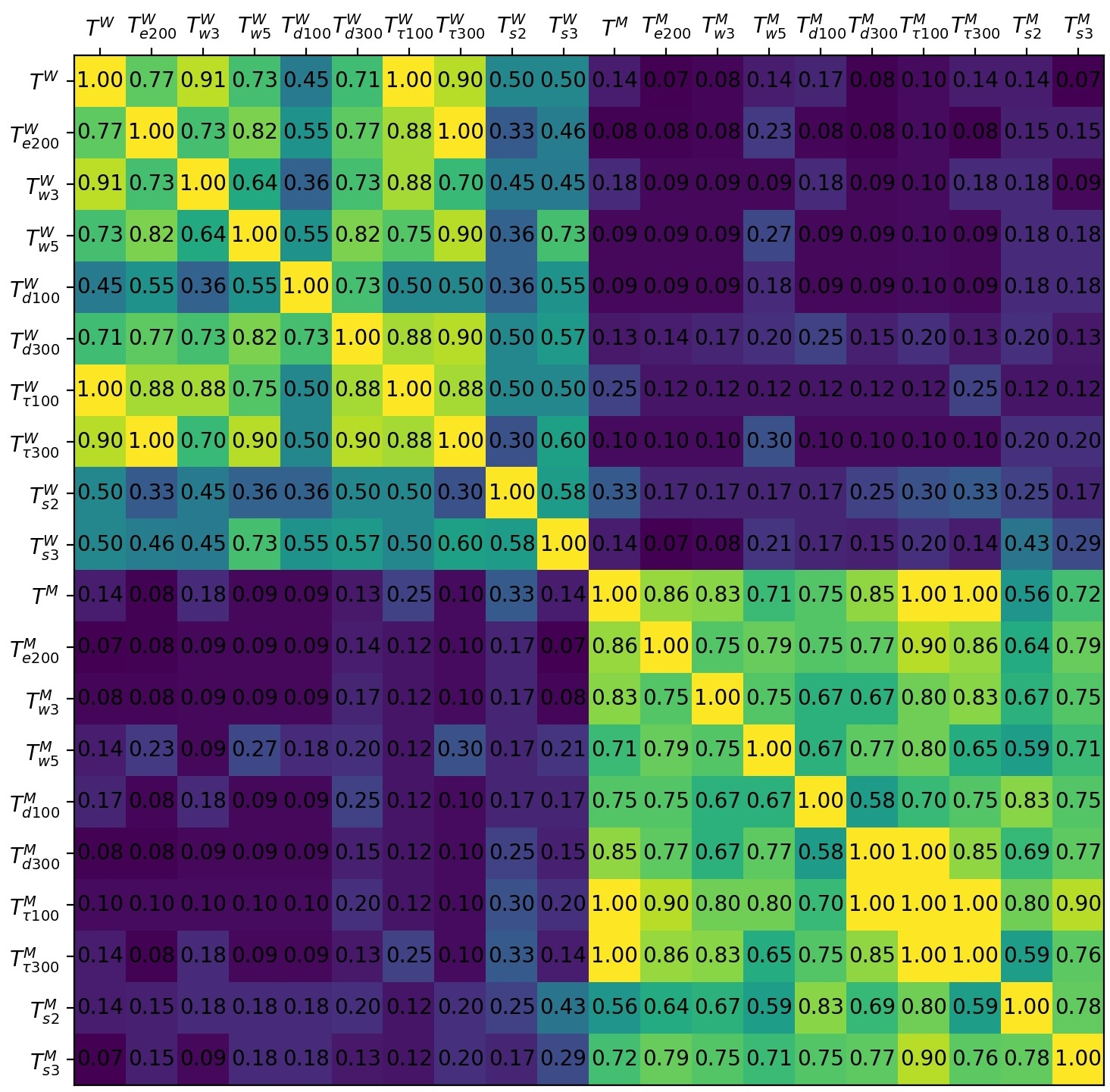}
  \label{fig:hm_dimensions_1}
  \vspace{-10pt}
\caption{Overlap coefficient between the set of biased concepts of the different models for /r/TheRedPill.}
\label{fig:hm_additional_results}
\end{figure}

We now provide a more detailed analysis on the effects that different training parameters have on the models created and the final set of discovered biased concepts to complement the experiments on stability presented above. For this, and due to a lack of space, we focus on the /r/TheRedPill community, which has the least number of unique words and the highest vocabulary density, so it is the most prone to suffer from stability issues that can cause differences in the biased concepts discovered given changes in the dataset and/or the parameters used.
In order to study the effect of all the parameters used, we trained five new models and ran nine new executions with different training and method parameters. 
When training the new models, only one parameter was changed at a time to study its influence \emph{ceteris paribus}. That is, all of the other parameters were set to the default values used to train the original /r/TheRedPill model $T$  (using a window size of 4, embedding dimension of 200, 100 epochs, $\tau=200$, and using a salience threshold of $n=4$ standard deviations  -- see Section \ref{sec:ev} for all the details). The specific ranges of the parameters used in the different executions ran are presented below:

\emph{Window sizes (w):} We trained two new models, $T_{w3}$ and $T_{w5}$, utilizing window sizes of 3 and 5 (instead of the window size of 4 used in the original models), to evaluate the effect of window sizes in the resulting set of discovered biased concepts.

\emph{Embedding dimension (d):} Models $T_{d100}$, and $T_{d300}$, were trained with 100 and 300 dimensions (instead of 200), respectively, to study the effect of embedding dimensionality.

\emph{Epochs (e):} Model $T_{e200}$ was trained using 200 epochs (instead of the 100 epochs of the original model), to study the effect of a longer training period. 

\emph{Cluster repetitions ($\tau$):} Models $T_{\tau100}$ and $T_{\tau300}$ were trained using the same training parameters as the original model, but with $\tau$ 100 and 300 cluster iterations, to select the partition with higher silhouette score in order to study its effect in the final set of biased concepts (instead of the $\tau=200$ used originally in Section~\ref{sec:ev}).

\emph{Salience thresholds (s):} Models $T_{s2}$ and $T_{s3}$ were trained with the same parameters as the original model, but using $n=2$ and $n=3$ standard deviations (instead of the $n=4$ used originally in Section~\ref{sec:ev}) to select the set of most salient words towards each attribute concept.

Figure \ref{fig:hm_additional_results} shows the overlapping coefficient between the sets of frequent biased concepts for women (represented with the superscript $W$) and men (represented with the superscript $M$) between the different models for /r/TheRedPill\footnote{Note that for the sake of clarity, we only considered labels used by more than 1\% of the clusters in each partition, selecting 14 labels for each model and attribute concept on average, in order to compare relatively meaningful and frequent biases in the community.} and the original model presented in Section \ref{sec:trp}, represented as $T$.
The results show that: i) an important overlap coefficient among the sets of biases towards the same gender for both women and men, indicating that biases towards women (and men) are similar among the different models, and 2) a strong difference between women and men biases in all models, indicating that the sets of discovered biased concepts towards the two genders are different in /r/TheRedPill.
Focusing on the influence of specific parameters, the results suggest that, among the different training parameters tested (training epochs, window sizes, and embedding dimensions), low embedding dimensionality ($d=100$) has the biggest impact when determining the set of biased concepts of a community, coinciding with previous studies that suggest there is a sweet spot for the dimensionality of word vectors \cite{melamud2016role}. Even in this case, the differences between women and men are still very marked. 
Importantly, and as one would expect, changing the $n$ for the salience threshold also affects the resulting sets of biased concepts as a consequence of clustering a different selection of salient words. Again, proportions between women and men are kept, but the more salient words considered ($n=2$) the more differences with the original. With lower values of $n$ we are including potentially less salient biases, so $n$ can act as a zoom in/out mechanism here as needed in the community of application, with higher values allowing to focus on the most salient (strong and frequent) biases of a community. Finally,
cluster repetitions $\tau$,  different window sizes ($w$) and epochs ($e$) seem to have little impact. 

\subsection{Validation}\label{sec:val}
Finally, to validate our method, we apply it to the large, widely-studied
Google News pre-trained model, and compare the results with predefined biases that had been \emph{attested} by previous work in this model~\cite{garg2018word,Caliskan2017}. The aim was to see whether our method would discover, among others, those predefined biases attested by prior work to validate that it finds relevant biases. 
The Google News model contains 300-dimensional vectors for 3 million words and phrases, trained on part of the US Google News dataset\footnote{\url{https://code.google.com/archive/p/word2vec/}}. Previous research on this model \emph{attested} the predefined gender biases and stereotypes
that associate women with \emph{family} and \emph{arts}, and men with \emph{career}, \emph{science} and \emph{maths} \cite{garg2018word,Caliskan2017}. 
We applied our method to the Google News pre-trained model, following the steps we describe in Section \ref{sec:main}. We first estimate the salience of the words in the vocabulary of the Google News model and select the most salient words towards \emph{women} and \emph{men}, creating $S1$ and $S2$ with sizes 1545 and 1271, respectively. The partition with higher silhouette score, using $\tau$= 200, clustered women-salient words in 508 clusters, and men-salient words in 552. Using WEAT to compare all women clusters with men clusters returns significant p-values for all comparison combinations, with p-values ranging between $1.5\cdot10^{-4}$ and $0.008$, passing the Benjamini-Hochberg correction test introduced before.

Here, we focus on the validation aspects, reporting whether our method also found, among all \emph{discovered} biases, the arbitrary biases that had been \emph{attested} by previous works.
Note that comparing both approaches is not straightforward, since \emph{discovering} the most biased words in a model is a different task than \emph{attesting} whether some arbitrary word vectors are strongly correlated: even when the attestation of an specific arbitrary bias results in significance, it may not be one of the strongest biased associations in the model and therefore not selected - among the other many discovered and stronger biased associations - by our approach. Our method discovers biased concepts towards two target sets, in this example \emph{men} and \emph{women}, but these most biased concepts do not necessarily have to be the arbitrary biases tested in previous works.  
An example of this situation can be seen when observing the Google News analysis in Appendix~\ref{ap:ap_gnews}: surprisingly no-one has previously attested biases related to \emph{physical appearance} in the Google News model, 
even though these biases are stronger and more pronounced (obtain more significant p-values when using WEAT) than the reported attested biased related to \emph{career, family, arts, science and maths} in previous work.

Therefore, to compare both approaches, we map the word sets used by previous works related to \emph{career, family, arts, science} and \emph{maths} to the most similar USAS semantic categories, based on an analysis of the WEAT word sets and the category descriptions provided in the USAS website\footnote{A list of all USAS labels can be accessed at \url{http://ucrel.lancs.ac.uk/usas/}}, and then count the number of clusters tagged with them for both target sets \emph{women} and \emph{men}. We consider that our method discovers similar biases attested by previous research if the associations reported in \cite{garg2018word,Caliskan2017} hold among the discovered biased concepts, that is, if we find more clusters tagged with labels related to \emph{career, maths} and \emph{science} among \emph{men}-biased concepts than among \emph{women}, and vice-versa for \emph{family} and \emph{arts}.

The association between the previously used WEAT word sets and USAS semantic categories goes as follows.
We map \emph{career words} to \emph{Money \& commerce in industry} and \emph{Power and organizing} USAS categories; \emph{family words} to \emph{Kin and People}; words related to \emph{arts} to \emph{Arts and Crafts}, words related to \emph{science} to \emph{Science and technology in general}, and \emph{mathematics} to the corresponding USAS label \emph{Mathematics}.

\begin{figure}[ht]
\centering
  \vspace{-5pt}
  \includegraphics[width=.8\linewidth]{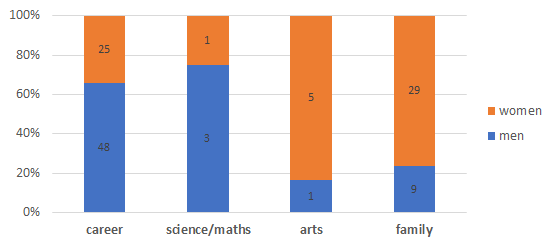}
  \vspace{-5pt}
\caption{Frequency of previously attested biased categories among discovered clusters in \emph{Google News}.}
\label{fig:strongbiases}
\end{figure}

Figure \ref{fig:strongbiases} shows that the semantic tags related to \emph{career} are strongly biased towards men, as identified in previous works, with double the number of clusters for men than women, containing words such as `boss' and `magnate' (for men)
and `chairwoman', `manageress', and `governess'  (for women).
\emph{Family}-related clusters are strongly biased towards women, with three times as many clusters for women than for men. Words include references to `mothering', `brides' and `niece' (women), and also `dad', `patriarch' and `nephew' (men). 
\textit{Arts} is also strongly biased towards women, with 5 times more clusters for women compared than for men, including words such as `knitting' and `crochet' for women, and `blacksmith' for men. 
Tags related to \emph{science} and \emph{maths} are not frequent among the set of most salient words in this model, but they are still more frequent among men-biased clusters than for women, with 3 and 1 clusters, respectively. 

This analysis shows that: i) our method \emph{discovers} relevant biases, as it discovers similar biases \emph{attested} by previous research; ii) as detailed in Appendix~\ref{ap:ap_gnews}, our method discovers additional biases related to gender in Google News not reported by previous work.

\section{Conclusions}\label{sec:conclusions}
In this paper, we improve and examine and validate a data-driven methodology for discovering biased concepts in linguistic corpora. 
Through the use of word embeddings and similarity metrics, which leverage the vocabulary used within specific communities, we are able to discover the biased concepts, in the form of clusters of words semantically related, towards different social groups when compared against each other. 
The resulting clusters abstract the inherent biases into more general (and stable) structures that allow a better understanding of the dispositions of these communities towards protected features.
 We discovered and analyzed gender and religion biases in two Reddit communities, and showed their stability. 
We also validated our method on the Google News dataset, confirming that it also discovers the arbitrary biases that had been attested by previous works. 

Quantifying language biases through our approach has academic and societal advantages. 
For socio-cultural linguistics, manual identification and analysis of the subcultures’ specialised vocabularies has become impractical; automatic methods that help with both the identification and understanding of community language. As a more general diagnostic, our method can help in understanding and measuring social problems and stereotypes towards certain populations in communities with more precision and clarity \cite{Abebe2019}. It can also promote discussions on the language used by particular communities. This is especially relevant considering the radicalisation of interest-based communities \cite{Marwick2017}. Our approach allows us to discover a broad categorical overview of biases, as well as a detailed analysis of these biases, as expressed in a community's own language. As such, we can trace language in cases where researchers do not know the specific linguistic forms employed in the community. This is all the more relevant given that online discourse communities are characterised by diachronic variability, with new users, topics and forms of dialect being introduced over time. As a bottom-up approach, our method can be used to monitor and account for these transformations. Finally, our method could underpin tools to help administrative bodies of web platforms to discover and trace biases in online communities to decide which do not conform to content policies.

It is finally important to note that, even though our method is automated, it is designed to be used with a human in the loop. 
This includes the need to consider in conjunction the more general and specific views provided by our method, and the wider context surrounding the biased concepts found, which is important to adequately interpret them. 
For instance, in /r/TheRedPill, the cluster \textit{casual} contains words that carry positive sentiment score, but these words are part of a discourse about having casual relationships with women in order to conquer as many of them as possible – objectifying them in the process. This is something that becomes clear when looking at the relative frequency of semantic categories discovered, with the most frequent ones including Relationship Intimate/Sexual, Appearance and Objects, among others. These findings do not offer conclusive evidence of stereotyping, but they can be valuable assets for the work of social scientists and industry experts who are dealing with language biases in their communities.
Finally, the target concepts should be defined with care, and the analysis based on them needs to be conducted with care as well. For instance, we considered gender as women vs men (as all the previous work related to gender biases in word embeddings to date did too). However, concepts may not always generalise in the same way and for every community, e.g., gender is increasingly considered fluid rather than binary \cite{Tannen1994, Witt2012}.

\section*{Code, Datasets and Tool}
To facilitate follow-up work, all the code and datasets are available \emph{\url{here}}.\footnote{\url{https://github.com/xfold/DiscoveringAndInterpretingConceptualBiases}}
The code repository has detailed instructions to replicate the experiments in the paper. 
Also, an interactive demo of our tool is \emph{here}.\footnote{\url{https://xfold.github.io/Web-DiscoveringAndInterpretingConceptualBiases/}}

\section*{Acknowledgments}
This research was funded by the EPSRC  under grant EP/R033188/1,  Discovering and Attesting Digital Discrimination (DADD) –  \url{https://dadd-project.github.io}.

\bibliographystyle{IEEEtran}
\bibliography{refs}

\begin{thebibliography}{10}
\providecommand{\url}[1]{#1}
\csname url@samestyle\endcsname
\providecommand{\newblock}{\relax}
\providecommand{\bibinfo}[2]{#2}
\providecommand{\BIBentrySTDinterwordspacing}{\spaceskip=0pt\relax}
\providecommand{\BIBentryALTinterwordstretchfactor}{4}
\providecommand{\BIBentryALTinterwordspacing}{\spaceskip=\fontdimen2\font plus
\BIBentryALTinterwordstretchfactor\fontdimen3\font minus
  \fontdimen4\font\relax}
\providecommand{\BIBforeignlanguage}[2]{{%
\expandafter\ifx\csname l@#1\endcsname\relax
\typeout{** WARNING: IEEEtran.bst: No hyphenation pattern has been}%
\typeout{** loaded for the language `#1'. Using the pattern for}%
\typeout{** the default language instead.}%
\else
\language=\csname l@#1\endcsname
\fi
#2}}
\providecommand{\BIBdecl}{\relax}
\BIBdecl

\bibitem{Hamilton1986}
D.~L. Hamilton and T.~K. Trolier, ``{Stereotypes and stereotyping: An overview
  of the cognitive approach.}'' in \emph{Prejudice, discrimination, and
  racism.}\hskip 1em plus 0.5em minus 0.4em\relax San Diego, CA, US: Academic
  Press, 1986, pp. 127--163.

\bibitem{Basow1992}
S.~A. Basow, \emph{Gender: Stereotypes and roles}.\hskip 1em plus 0.5em minus
  0.4em\relax Thomson Brooks/Cole Publishing Co, 1992.

\bibitem{Holmes2008}
J.~Holmes and M.~Meyerhoff, \emph{{The handbook of language and gender}}.\hskip
  1em plus 0.5em minus 0.4em\relax Hoboken: John Wiley {\&} Sons, 2008,
  vol.~25.

\bibitem{bolukbasi2016man}
T.~Bolukbasi, K.-W. Chang, J.~Y. Zou, V.~Saligrama, and A.~T. Kalai, ``Man is
  to computer programmer as woman is to homemaker? debiasing word embeddings,''
  in \emph{Advances in Neural Information Processing Systems}, 2016, pp.
  4349--4357.

\bibitem{Caliskan2017}
A.~Caliskan, J.~J. Bryson, and A.~Narayanan, ``{Semantics derived automatically
  from language corpora contain human-like biases},'' \emph{Science}, vol. 356,
  no. 6334, pp. 183--186, 2017.

\bibitem{barocas2017problem}
S.~Barocas, K.~Crawford, A.~Shapiro, and H.~Wallach, ``The problem with bias:
  from allocative to representational harms in machine learning. special
  interest group for computing,'' \emph{Information and Society (SIGCIS)},
  vol.~2, 2017.

\bibitem{zemel2013learning}
R.~Zemel, Y.~Wu, K.~Swersky, T.~Pitassi, and C.~Dwork, ``Learning fair
  representations,'' in \emph{International Conference on Machine Learning},
  2013, pp. 325--333.

\bibitem{criadodigital}
N.~Criado and J.~Such, ``Digital discrimination,'' in \emph{Algorithmic
  Regulation}.\hskip 1em plus 0.5em minus 0.4em\relax Oxford University Press,
  pp. 82--97.

\bibitem{van2020transparency}
T.~van Nuenen, X.~Ferrer, J.~Such, and M.~Cot{\'e}, ``Transparency for whom?
  assessing discriminatory artificial intelligence,'' \emph{Computer}, vol.~53,
  no.~11, pp. 36--44, 2020.

\bibitem{ferrer2020bias}
X.~F. Aran, T.~van Nuenen, J.~Such, M.~Cot{\'e}, and N.~Criado, ``Bias and
  discrimination in ai: a cross-disciplinary perspective,'' \emph{IEEE
  Technology and Society}, vol.~40, no.~2, pp. 72--80, 2021.

\bibitem{such2017privacy}
J.~Such, ``Privacy and autonomous systems,'' in \emph{Proceedings of the
  Twenty-Sixth International Joint Conference on Artificial Intelligence,
  IJCAI-17}, 2017, pp. 4761--4767.

\bibitem{Bryson2016}
V.~Bryson, \emph{{Feminist political theory}}.\hskip 1em plus 0.5em minus
  0.4em\relax Macmillan International Higher Education, 2016.

\bibitem{Zhao2006}
B.~Zhao, J.~Ondrich, and J.~Yinger, ``{Why do real estate brokers continue to
  discriminate? Evidence from the 2000 Housing Discrimination Study},''
  \emph{Journal of Urban Economics}, vol.~59, no.~3, pp. 394--419, 2006.

\bibitem{VanMiltenburg2016}
C.~van Miltenburg, ``Stereotyping and bias in the flickr30k dataset,'' in
  \emph{11th workshop on multimodal corpora: computer vision and language
  processing}, 2016.

\bibitem{garg2018word}
N.~Garg, L.~Schiebinger, D.~Jurafsky, and J.~Zou, ``Word embeddings quantify
  100 years of gender and ethnic stereotypes,'' \emph{PNAS 2018}, vol. 115,
  no.~16, pp. E3635--E3644, 2018.

\bibitem{kozlowski2019geometry}
A.~C. Kozlowski, M.~Taddy, and J.~A. Evans, ``The geometry of culture:
  Analyzing the meanings of class through word embeddings,'' \emph{American
  Sociological Review}, vol.~84, no.~5, pp. 905--949, 2019.

\bibitem{Fine1979}
\BIBentryALTinterwordspacing
G.~A. Fine and S.~Kleinman, ``{Rethinking Subculture: An Interactionist
  Analysis},'' \emph{American Journal of Sociology}, vol.~85, no.~1, pp. 1--20,
  aug 1979. [Online]. Available: \url{http://www.jstor.org/stable/2778065}
\BIBentrySTDinterwordspacing

\bibitem{Marwick2017}
A.~Marwick and R.~Lewis, ``Media manipulation and disinformation online,''
  \emph{New York: Data \& Society Research Institute}, pp. 7--19, 2017.

\bibitem{Abebe2019}
R.~Abebe, S.~Barocas, J.~Kleinberg, K.~Levy, M.~Raghavan, and D.~G. Robinson,
  ``Roles for computing in social change,'' in \emph{Procs. of the Conference
  on Fairness, Accountability, and Transparency (FAT)}, 2020, pp. 252--260.

\bibitem{tanczer2016hacktivism}
L.~M. Tanczer, ``Hacktivism and the male-only stereotype,'' \emph{New Media \&
  Society}, vol.~18, no.~8, pp. 1599--1615, 2016.

\bibitem{creese2019stereotypes}
A.~Creese and A.~Blackledge, ``Stereotypes and chronotopes: The peasant and the
  cosmopolitan in narratives about migration,'' \emph{Journal of
  Sociolinguistics}, 2019.

\bibitem{aran2019attesting}
X.~F. Aran, J.~Such, and N.~Criado, ``Attesting biases and discrimination using
  language semantics,'' in \emph{Responsible Artificial Intelligence Agents
  workshop of the International Conference on Autonomous Agents and Multiagent
  Systems (AAMAS 2019)}, 2019.

\bibitem{ferrer2020discovering}
X.~Ferrer, T.~van Nuenen, J.~Such, and N.~Criado, ``Discovering and
  categorising language biases in reddit,'' in \emph{Proceedings of the
  International AAAI Conference on Web and Social Media}, vol.~15, no.~1, 2021,
  pp. 140--151.

\bibitem{sutton2018biased}
A.~Sutton, T.~Lansdall-Welfare, and N.~Cristianini, ``Biased embeddings from
  wild data: Measuring, understanding and removing,'' in \emph{International
  Symposium on Intelligent Data Analysis}.\hskip 1em plus 0.5em minus
  0.4em\relax Springer, 2018, pp. 328--339.

\bibitem{zhao2019gender}
J.~Zhao, T.~Wang, M.~Yatskar, R.~Cotterell, V.~Ordonez, and K.-W. Chang,
  ``Gender bias in contextualized word embeddings,'' in \emph{Procs. of the
  Conference of the North American Chapter of the Association for Computational
  Linguistics)}, 2019, pp. 629--634.

\bibitem{papakyriakopoulos2020bias}
O.~Papakyriakopoulos, S.~Hegelich, J.~C.~M. Serrano, and F.~Marco, ``Bias in
  word embeddings,'' in \emph{Proceedings of the 2020 Conference on Fairness,
  Accountability, and Transparency}, 2020, pp. 446--457.

\bibitem{manzini2019black}
T.~Manzini, L.~Y. Chong, A.~W. Black, and Y.~Tsvetkov, ``Black is to criminal
  as caucasian is to police: Detecting and removing multiclass bias in word
  embeddings,'' in \emph{Procs. of the Conf. of the North American Chapter of
  the Association for Comp. Linguistics}, 2019, pp. 615--621.

\bibitem{kurita2019measuring}
K.~Kurita, N.~Vyas, A.~Pareek, A.~W. Black, and Y.~Tsvetkov, ``Measuring bias
  in contextualized word representations,'' in \emph{Procs. of the Workshop on
  Gender Bias in Natural Language Processing}, 2019, pp. 166--172.

\bibitem{brunet2018understanding}
M.-E. Brunet, C.~Alkalay-Houlihan, A.~Anderson, and R.~Zemel, ``Understanding
  the origins of bias in word embeddings,'' in \emph{International Conference
  on Machine Learning}, 2019, pp. 803--811.

\bibitem{greenwald1998measuring}
A.~G. Greenwald, D.~E. McGhee, and J.~L. Schwartz, ``Measuring individual
  differences in implicit cognition: the implicit association test.'' \emph{J
  Pers Soc Psychol}, vol.~74, no.~6, p. 1464, 1998.

\bibitem{kiefer2007implicit}
A.~K. Kiefer and D.~Sekaquaptewa, ``Implicit stereotypes and women’s math
  performance: How implicit gender-math stereotypes influence women’s
  susceptibility to stereotype threat,'' \emph{Journal of experimental social
  psychology}, vol.~43, no.~5, pp. 825--832, 2007.

\bibitem{zhang2018mitigating}
B.~H. Zhang, B.~Lemoine, and M.~Mitchell, ``Mitigating unwanted biases with
  adversarial learning,'' in \emph{Proceedings of the 2018 AAAI/ACM Conference
  on AI, Ethics, and Society}, 2018, pp. 335--340.

\bibitem{swinger2019biases}
N.~Swinger, M.~De-Arteaga, N.~T. Heffernan~IV, M.~D. Leiserson, and A.~T.
  Kalai, ``What are the biases in my word embedding?'' in \emph{Proceedings of
  the 2019 AAAI/ACM Conference on AI, Ethics, and Society}, 2019, pp. 305--311.

\bibitem{gonen2019lipstick}
H.~Gonen and Y.~Goldberg, ``Lipstick on a pig: Debiasing methods cover up
  systematic gender biases in word embeddings but do not remove them,'' in
  \emph{Procs. of the 2019 Conference of the North American Chapter of the
  Association for Computational Linguistics}, 2019, pp. 609--614.

\bibitem{bodell2019interpretable}
M.~H. Bodell, M.~Arvidsson, and M.~Magnusson, ``Interpretable word embeddings
  via informative priors,'' in \emph{Proceedings of the 2019 Conference on
  Empirical Methods in Natural Language Processing and the 9th International
  Joint Conference on Natural Language Processing (EMNLP-IJCNLP)}, 2019, pp.
  6324--6330.

\bibitem{Farrell2020}
\BIBentryALTinterwordspacing
T.~Farrell, O.~Araque, M.~Fernandez, and H.~Alani, ``On the use of jargon and
  word embeddings to explore subculture within the reddit’s manosphere,'' in
  \emph{12th ACM Conference on Web Science}, ser. WebSci '20.\hskip 1em plus
  0.5em minus 0.4em\relax New York, NY, USA: Association for Computing
  Machinery, 2020, p. 221–230. [Online]. Available:
  \url{https://doi.org/10.1145/3394231.3397912}
\BIBentrySTDinterwordspacing

\bibitem{MacQueen2001}
\BIBentryALTinterwordspacing
K.~M. MacQueen, E.~McLellan, D.~S. Metzger, S.~Kegeles, R.~P. Strauss,
  R.~Scotti, L.~Blanchard, and R.~T. {Trotter 2nd},
  ``\BIBforeignlanguage{eng}{{What is community? An evidence-based definition
  for participatory public health}},'' \emph{\BIBforeignlanguage{eng}{American
  journal of public health}}, vol.~91, no.~12, pp. 1929--1938, dec 2001.
  [Online]. Available: \url{https://pubmed.ncbi.nlm.nih.gov/11726368
  https://www.ncbi.nlm.nih.gov/pmc/articles/PMC1446907/}
\BIBentrySTDinterwordspacing

\bibitem{badillawefe}
P.~Badilla, F.~Bravo-Marquez, and J.~P{\'e}rez, ``Wefe: The word embeddings
  fairness evaluation framework,'' in \emph{International Joint Conference on
  Artificial Intelligence (IJCAI)}, 2020.

\bibitem{Antoniak2018}
M.~Antoniak and D.~Mimno, ``{Evaluating the Stability of Embedding-based Word
  Similarities},'' \emph{Transactions of the Association for Computational
  Linguistics}, vol.~6, pp. 107--119, 2018.

\bibitem{newman2005power}
M.~E. Newman, ``Power laws, pareto distributions and zipf's law,''
  \emph{Contemporary physics}, vol.~46, no.~5, pp. 323--351, 2005.

\bibitem{rousseeuw1987silhouettes}
P.~J. Rousseeuw, ``Silhouettes: a graphical aid to the interpretation and
  validation of cluster analysis,'' \emph{Journal of computational and applied
  mathematics}, vol.~20, pp. 53--65, 1987.

\bibitem{hamerly2004learning}
G.~Hamerly and C.~Elkan, ``Learning the k in k-means,'' in \emph{Advances in
  neural information processing systems}, 2004, pp. 281--288.

\bibitem{garg2018supervising}
V.~Garg and A.~T. Kalai, ``Supervising unsupervised learning,'' in
  \emph{Advances in Neural Information Processing Systems}, 2018, pp.
  4991--5001.

\bibitem{benjamini1995controlling}
Y.~Benjamini and Y.~Hochberg, ``Controlling the false discovery rate: a
  practical and powerful approach to multiple testing,'' \emph{Journal of the
  Royal statistical society: series B (Methodological)}, vol.~57, no.~1, pp.
  289--300, 1995.

\bibitem{diz2011multiple}
A.~P. Diz, A.~Carvajal-Rodr{\'\i}guez, and D.~O. Skibinski, ``Multiple
  hypothesis testing in proteomics: a strategy for experimental work,''
  \emph{Molecular \& Cellular Proteomics}, vol.~10, no.~3, 2011.

\bibitem{rayson2004ucrel}
P.~Rayson, D.~Archer, S.~Piao, and A.~M. McEnery, ``The ucrel semantic analysis
  system.'' in \emph{In Procs of the workshop on Beyond Named Entity
  Recognition Semantic labelling for NLP tasks, 4th Int. Conf. on Language
  Resources and Evaluation (LREC)}, 2004, pp. 7--12.

\bibitem{wilson1993automatic}
A.~Wilson and P.~Rayson, ``Automatic content analysis of spoken discourse: a
  report on work in progress,'' \emph{Corpus based computational linguistics},
  pp. 215--226, 1993.

\bibitem{sharoff2006assist}
S.~Sharoff, B.~Babych, P.~Rayson, O.~Mudraya, and S.~Piao, ``Assist: Automated
  semantic assistance for translators,'' in \emph{Proceedings of the Eleventh
  Conference of the European Chapter of the Association for Computational
  Linguistics}, 2006, pp. 139--142.

\bibitem{rayson2008key}
P.~Rayson, ``From key words to key semantic domains,'' \emph{International
  journal of corpus linguistics}, vol.~13, no.~4, pp. 519--549, 2008.

\bibitem{feldman2013techniques}
R.~Feldman, ``Techniques and applications for sentiment analysis,''
  \emph{Communications of the ACM}, vol.~56, no.~4, pp. 82--89, 2013.

\bibitem{Watson2016}
\BIBentryALTinterwordspacing
Z.~Watson, ``{Red Pill Men and Women, Reddit, And The Cult of Gender |
  Inverse},'' 2016. [Online]. Available:
  \url{https://www.inverse.com/article/15832-red-pill-men-and-women-reddit-and-the-cult-of-gender}
\BIBentrySTDinterwordspacing

\bibitem{nosek2002harvesting}
B.~A. Nosek, M.~R. Banaji, and A.~G. Greenwald, ``Harvesting implicit group
  attitudes and beliefs from a demonstration web site.'' \emph{Group Dynamics:
  Theory, Research, and Practice}, vol.~6, no.~1, p. 101, 2002.

\bibitem{Baumgartner2020}
J.~Baumgartner, S.~Zannettou, B.~Keegan, M.~Squire, and J.~Blackburn, ``The
  pushshift reddit dataset,'' in \emph{Proceedings of the international AAAI
  conference on web and social media}, vol.~14, 2020, pp. 830--839.

\bibitem{peters2018deep}
M.~E. Peters, M.~Neumann, M.~Iyyer, M.~Gardner, C.~Clark, K.~Lee, and
  L.~Zettlemoyer, ``Deep contextualized word representations,'' in
  \emph{Proceedings of NAACL-HLT}, 2018, pp. 2227--2237.

\bibitem{devlin2018bert}
J.~Devlin, M.-W. Chang, K.~Lee, and K.~Toutanova, ``Bert: Pre-training of deep
  bidirectional transformers for language understanding,'' \emph{arXiv preprint
  arXiv:1810.04805}, 2018.

\bibitem{hutto2014vader}
C.~J. Hutto and E.~Gilbert, ``Vader: A parsimonious rule-based model for
  sentiment analysis of social media text,'' in \emph{Eighth international AAAI
  conference on weblogs and social media}, 2014.

\bibitem{LaViolette2019}
J.~LaViolette and B.~Hogan, ``{Using platform signals for distinguishing
  discourses: The case of men's rights and men's liberation on Reddit},''
  \emph{ICWSM 2019}, pp. 323--334, 2019.

\bibitem{Schmitz2016}
R.~Schmitz and E.~Kazyak, ``{Masculinities in Cyberspace: An Analysis of
  Portrayals of Manhood in Men's Rights Activist Websites},'' \emph{Social
  Sciences}, vol.~5, no.~2, p.~18, 2016.

\bibitem{Ging2017}
D.~Ging, ``{Alphas, Betas, and Incels: Theorizing the Masculinities of the
  Manosphere},'' \emph{Men and Masculinities}, vol.~22, no.~4, pp. 638--657,
  2019.

\bibitem{Mountford2018}
J.~Mountford, ``{Topic Modeling The Red Pill},'' \emph{Social Sciences},
  vol.~7, no.~3, p.~42, 2018.

\bibitem{Ribeiro2020b}
\BIBentryALTinterwordspacing
M.~H. Ribeiro, J.~Blackburn, B.~Bradlyn, E.~{De Cristofaro}, G.~Stringhini,
  S.~Long, S.~Greenberg, and S.~Zannettou, ``{From Pick-Up Artists to Incels: A
  Data-Driven Sketch of the Manosphere},'' 2020. [Online]. Available:
  \url{http://arxiv.org/abs/2001.07600}
\BIBentrySTDinterwordspacing

\bibitem{shiebler2018fighting}
D.~Shiebler, L.~Belli, J.~Baxter, H.~Xiong, and A.~Tayal, ``Fighting redundancy
  and model decay with embeddings,'' \emph{arXiv preprint arXiv:1809.07703},
  2018.

\bibitem{vijaymeena2016survey}
M.~Vijaymeena and K.~Kavitha, ``A survey on similarity measures in text
  mining,'' \emph{Machine Learning and Applications: An International Journal},
  vol.~3, no.~2, pp. 19--28, 2016.

\bibitem{melamud2016role}
O.~Melamud, D.~McClosky, S.~Patwardhan, and M.~Bansal, ``The role of context
  types and dimensionality in learning word embeddings,'' \emph{arXiv preprint
  arXiv:1601.00893}, 2016.

\bibitem{Tannen1994}
D.~Tannen, ``{Gender and Discourse},'' Oxford and New York, 1994.

\bibitem{Witt2012}
C.~Witt, \emph{{The Metaphysics of Gender}}.\hskip 1em plus 0.5em minus
  0.4em\relax Oxford and New York: Oxford University Press, 2012.

\end{thebibliography}

\vfill
\newpage
\clearpage

\appendices

\section{Google News Analysis}\label{ap:ap_gnews}
\begin{figure}[h]
  \centering
 \vspace*{-5pt}
  \includegraphics[width=0.9\linewidth]{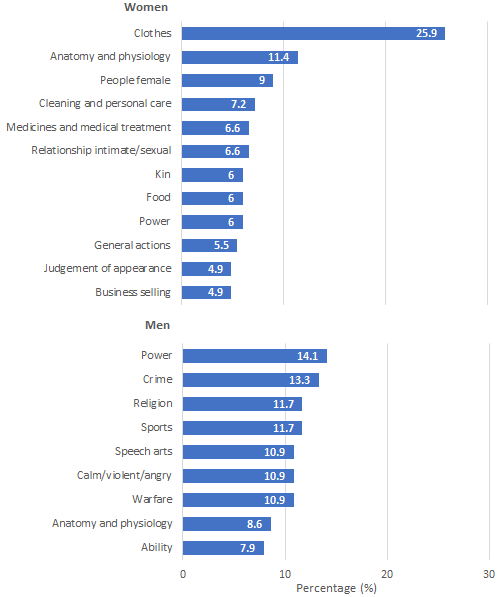}
  \caption{Relative frequency of semantic categories for biased concepts towards \emph{Women} (top) and \emph{Men} (bottom) in Google News.}
  \label{fig:gnews_m1}
\end{figure}

\small
Figure \ref{fig:gnews_m1} shows the general distribution of biased concepts among all clusters for towards \emph{women} and \emph{men}, respectively, showing only labels most frequent than 1\% of the total label frequency, and ignoring those clusters without a conceptual label for both women (273) and men (260). 
The general analysis of the community's biased concepts shows that some of the most common biases towards \emph{women} in Google News are related with physical appearance (such as Clothes, Anatomy and Physiology, Personal Care, and Judgment of Appearance) and relationships. On the other hand, the most frequent men-biased labels are related with Power, Violence, Religion and sports. 

\begin{figure}[ht!]
\centering
  \includegraphics[width=.8\linewidth]{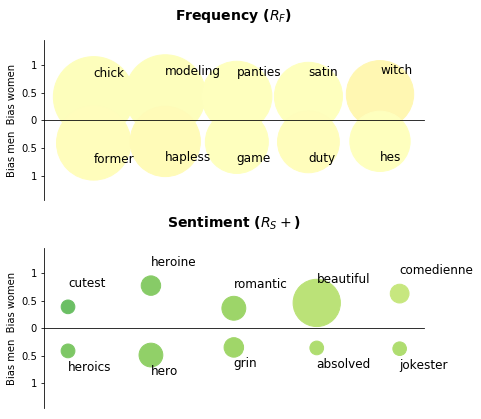}
  \caption{Top-5 $R_f$ and $R_{S}+$ clusters in GoogleNews.}
  \label{fig:gnews1}
\end{figure}

\begin{figure}[ht!]
\centering
  \vspace*{-5mm}
  \includegraphics[width=.8\linewidth]{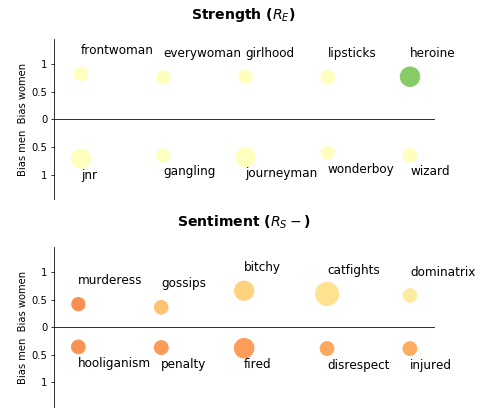}
  \caption{Top-5 $R_E$ and $R_{S}-$ clusters in GoogleNews.}
  \label{fig:gnews2}
\end{figure}

Figures \ref{fig:gnews1} and \ref{fig:gnews2} show the top 5 clusters for \emph{women} and \emph{men} in Google News by rank. Women-biased clusters are shown on top and men-biased clusters on the bottom, the y-axis corresponds with average salience of the cluster, size corresponds with frequency, and colour with average sentiment. 
Women-biased clusters are often related with physical appearance (for instance clusters `modelling', `lipsticks', `beautiful', or `cutest'), while men-biased clusters are often related with leadership, sports and strength (e.g. `hero', `duty', or `wonderboy'). 
This picture is clear if we focus in the top most frequent biased clusters. For \emph{women} we find words such as `gal', `sweetie' and `teeny' (grouped in `chick' cluster), `curvy' and `voluptuous' in the `modelling' cluster,  and words such as `pastel', `fuchsia', `skanky' and `slut' inside  `satin' and `witch' clusters, respectively. For \emph{men}, we find words such as 
`general', `veteran', `pro' and `astute' (in `former' cluster), `bumbling', `weakling' and `slacker' in `hapless' cluster, and words related to sports and strength such as `squad', `pitch', `fraternity' and `skillfull' in the other clusters shown in Figure \ref{fig:gnews2}.
Although not as strongly biased as in /r/TheRedPill, in Google News we find clusters with obvious negative connotations towards women such as `witch', `bitchy', `gossips' and `murderess', among the most frequent and negative women-biased rankings, respectively. On the other hand, the most negative clusters towards men are related to sports, being `hooliganism', `penalty', and `disrespect' some of the most negative. 
Finally, the most positive rankings show that most of the \emph{women} biased words are, again and excluding `heroine' cluster, related to physical appearance: i.e. `cutest', `beautiful', `gorgeous' and `ravishing'. For \emph{men} some of the most positive clusters contain words such as `heroics', `heroes', `absolved' and `jokester'.\footnote{Note that all details of the biases discovered and biased concepts are publicly available in our Github repository, which is accessible at \url{https://github.com/xfold/DiscoveringAndInterpretingConceptualBiases}.}


\section{Attribute and Other Sets}\label{ap:targetsets}

\noindent\textbf{/r/TheRedPill attribute concepts}
From \cite{nosek2002harvesting}.
{
 \emph{Female (women)}: sister, female, woman, girl, daughter, she, hers, her. \emph{Male (men)}: brother, male, man, boy, son, he, his, him.
 }

\noindent\textbf{/r/Atheism attribute concepts}\label{sec:atheismsets}
From \cite{garg2018word}.
{
 \emph{Islam words:} allah, ramadan, turban, emir, salaam, sunni, koran, imam, sultan, prophet, veil, ayatollah, shiite, mosque, islam, sheik, muslim, muhammad. \emph{Christianity words}: baptism, messiah, catholicism, resurrection, christianity, salvation, protestant, gospel, trinity, jesus, christ, christian, cross, catholic, church
 }

\noindent\textbf{Google News attribute and target sets}\label{sec:gnewssets}
From \cite{garg2018word}.
{
\emph{Women}: sister, female, woman, girl, daughter, she, hers, her. \emph{Men}: brother, male, man, boy, son, he, his, him. \emph{Career words}: executive, management, professional, corporation, salary, office, business, career. \emph{Family}: home, parents, children, family, cousins, marriage, wedding, relatives. \emph{Math}: math, algebra, geometry, calculus, equations, computation, numbers, addition. \emph{Arts}: poetry, art, sculpture, dance, literature, novel, symphony, drama. \emph{Science}: science, technology, physics  , chemistry, Einstein, NASA, experiment, astronomy.
}

\end{document}